\documentclass[10pt,twocolumn,letterpaper]{article}

\usepackage[]{iccv}      
\usepackage{multirow}
\usepackage{adjustbox}
\usepackage{float}
\usepackage{stfloats}

\definecolor{iccvblue}{rgb}{0.21,0.49,0.74}
\usepackage[pagebackref,breaklinks,colorlinks,allcolors=iccvblue]{hyperref}
\def\paperID{14826} 
\def\confName{ICCV}
\def\confYear{2025}

\title{MaskAttn-UNet: A Mask Attention-Driven Framework for Universal Low-Resolution Image Segmentation}

\author{%
Anzhe Cheng$^{1}$, Chenzhong Yin$^{1}$, Yu Chang$^{2}$, Heng Ping$^{1}$, Shixuan Li$^{1}$, Shahin Nazarian$^{1}$, Paul Bogdan$^{1}$\\[1ex]
$^{1}$University of Southern California\\
$^{2}$ The University of British Columbia\\[1ex]
}

\begin{document}
\maketitle
\begin{abstract}
Low-resolution image segmentation is crucial in real-world applications such as robotics, augmented reality, and large-scale scene understanding, where high-resolution data is often unavailable due to computational constraints. To address this challenge, we propose \textbf{MaskAttn-UNet}, a novel segmentation framework that enhances the traditional U-Net architecture via a mask attention mechanism. Our model selectively emphasizes important regions while suppressing irrelevant backgrounds, thereby improving segmentation accuracy in cluttered and complex scenes. Unlike conventional U-Net variants, MaskAttn-UNet effectively balances local feature extraction with broader contextual awareness, making it particularly well-suited for low-resolution inputs. We evaluate our approach on three benchmark datasets with input images rescaled to $128\times128$ and demonstrate competitive performance across semantic, instance, and panoptic segmentation tasks. Our results show that MaskAttn-UNet achieves accuracy comparable to state-of-the-art methods at significantly lower computational cost than transformer-based models, making it an efficient and scalable solution for low-resolution segmentation in resource-constrained scenarios.
\end{abstract}    
\begin{figure*}
    \centering
    \includegraphics[width=\linewidth]{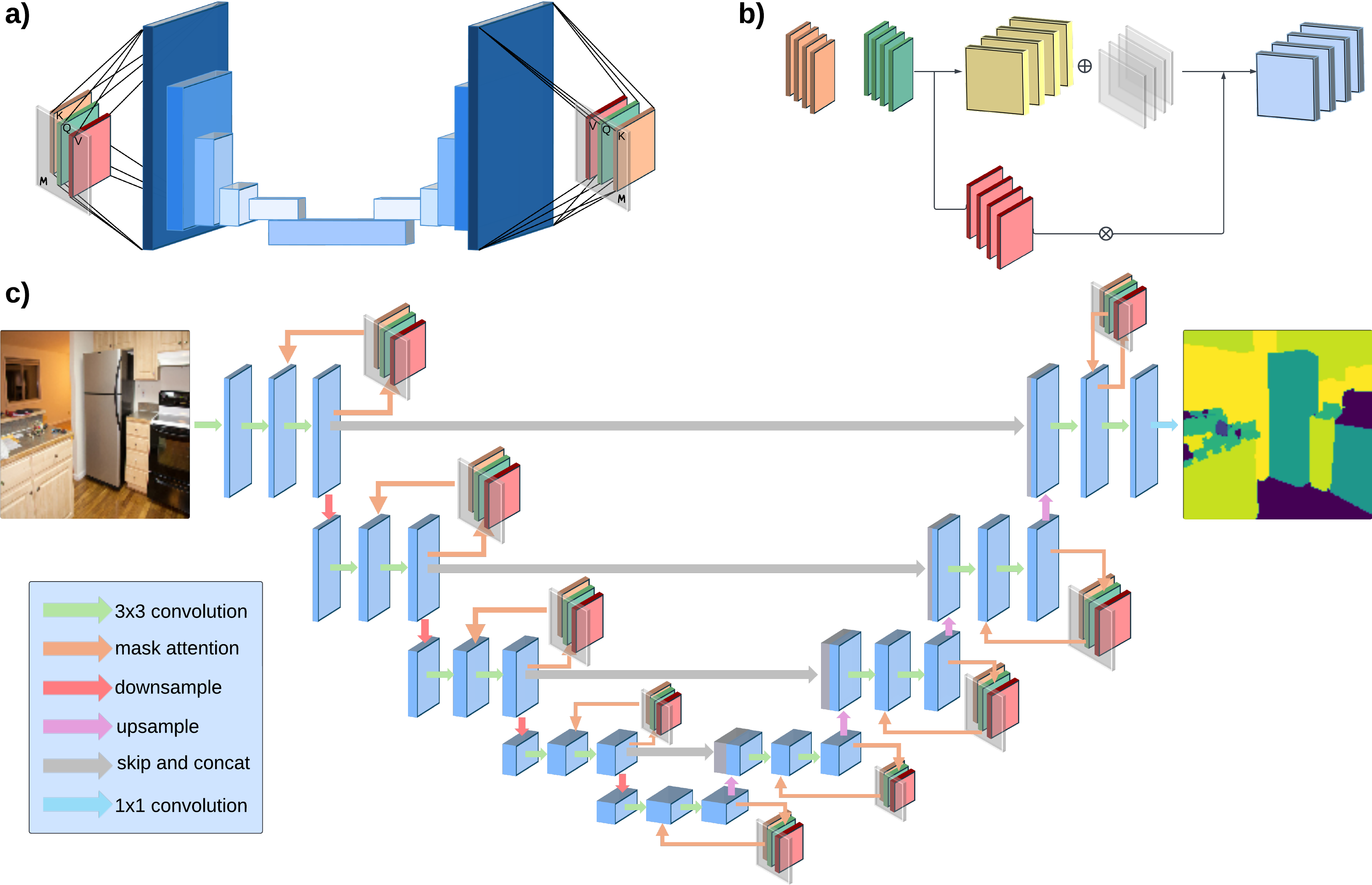}
    \caption{\textbf{Overview of the proposed MaskAttn-UNet.} (a) Overall architecture with a U-Net encoder-decoder and skip connections. (b) Mask Attention Module applying a learnable mask to modulate self-attention. (c) Multi-scale encoder–decoder design with convolutional layers, mask attention at each scale and skip connections between encoder and decoder.}
    \label{fig:overview}
    \vspace{-3mm}
\end{figure*}
\section{Introduction}
\label{sec:intro}
Accurate multi-class segmentation in complex scenes is crucial for applications such as autonomous driving, robotics, and augmented reality~\cite{minaee2021image,yao2023radar,almujally2024unet}. In autonomous vehicles, for example, precise pixel-wise labeling of vehicles and pedestrians is essential for safe navigation~\cite{bhatia2023road,chen2022gocomfort}, while in industrial robotics, detailed segmentation of tools and obstacles enables reliable manipulation and effective obstacle avoidance~\cite{kabir2025terrain,thakur2024depth}. However, many practical vision systems — from low-cost surveillance cameras to unmanned aerial vehicles (UAVs) and mobile robots — operate on low-resolution imagery due to sensor constraints and hardware limitations~\cite{xiang2019mini,joshi2024detection,aakerberg2022real}. This reduction in image detail poses a significant challenge for segmentation algorithms~\cite{litjens2014evaluation, bhanu1995adaptive, ghosh2019understanding, minaee2021image}, which must still accurately distinguish multiple object classes. Consequently, there is a pressing need for segmentation methods that remain robust and precise under such constrained conditions.

Encoder-decoder architectures like U-Net~\cite{ronneberger2015u} have proven effective at extracting local features and fine details through their multi-scale architecture. Nevertheless, they often struggle to capture long-range dependencies when multiple objects or classes coexist in a single image~\cite{chen2025u,zhang2024weakly}, leading to ambiguities in complex scenes~\cite{song2024multi,rmr2024multi}.
Conversely, transformer-based vision models incorporate global context through self-attention mechanisms, enabling them to model long-range relationships between pixels or regions~\cite{yang2021focal,shi2023spatial}. This global representation comes at the cost of substantial memory and computation overhead due to the quadratic complexity of self-attention, which can render such models impractical for embedded or real-time systems~\cite{feeley1995implementing,chen2016training}. Additionally, because vision transformers lack the inherent inductive biases of CNNs (especially the locality bias), fully attention-driven models may overlook the fine-grained details needed to distinguish small or overlapping objects~\cite{bera2021attend,rekavandi2023transformers,xu2021vitae}. These limitations highlight the need for a segmentation approach that balances local feature precision, global context capture, and computational efficiency.

In this paper, we introduce \textbf{MaskAttn-UNet}, an innovative extension of the U-Net framework that integrates a novel mask attention module to address the above challenges.
The MaskAttn-UNet architecture preserves U-Net’s strength in capturing fine local details via its skip connections, while the mask attention module selectively emphasizes salient regions in feature maps to inject broader contextual information. By focusing attention on relevant regions (instead of attending globally to all pixels), our approach can capture long-range dependencies more efficiently and mitigate the memory burden typically associated with transformers.
We specifically design the network for low-resolution inputs ($128\times128$ images), which significantly reduces computational demands while still allowing the model to learn rich representations. This design choice reflects real-world use cases with limited image resolutions and ensures that MaskAttn-UNet remains suitable for resource-constrained schemes.

We evaluate MaskAttn-UNet on standard benchmarks across semantic, instance, and panoptic segmentation tasks. Despite operating on relatively low-resolution inputs, our proposed model achieves competitive performance in terms of mean Intersection-over-Union (mIoU), Panoptic Quality (PQ), and Average Precision (AP) compared to state-of-the-art methods. Notably, MaskAttn-UNet maintains a moderate memory footprint during inference, making it significantly more practical for deployment than many fully transformer-based models that offer similar accuracy. These results demonstrate that our hybrid approach effectively combines the benefits of convolutional inductive bias and targeted self-attention, yielding robust multi-class segmentation in diverse and complex scenes.

Our contributions are summarized as follows:
\begin{itemize}
    \item We propose MaskAttn-UNet, a self-attention U-Net variant that integrates a novel mask attention module to capture both local details and long-range dependencies effectively.
    \item We design the architecture for low-resolution segmentation using $128\times128$ inputs, reducing computational demands while preserving robust performance.
    \item We validate our approach on several datasets, demonstrating improvements in segmentation metrics with lower memory consumption relative to transformer-based methods.
\end{itemize}
\vspace{-0.7mm}
Together, these contributions highlight the value of combining convolutional inductive biases with targeted attention mechanisms to achieve accurate and efficient segmentation in real-world plots. In the following sections, we discuss related work that motivated our approach, including U-Net extensions, vision transformers, and mask-based segmentation methods.
\section{Related Work}
\label{sec:related}

\subsection{U-Net}
\vspace{-2mm}
U-Net~\cite{ronneberger2015u} introduced an encoder-decoder architecture that has become a cornerstone in image segmentation. Its design consists of a contracting path that employs successive convolutions and pooling operations to extract features at multiple scales and an expansive path that uses upsampling layers to recover spatial resolution. Skip connections between corresponding layers in the encoder and decoder allow the network to merge deep semantic information with high-resolution spatial details. This structure has proven effective in applications such as biomedical segmentation, where precise localization is critical~\cite{azad2024medical, weng2021inet,ronneberger2015u,huang2024multiscaleconditionalgenerativemodeling}.

Despite its success, the fixed receptive fields inherent in standard convolutional layers restrict U-Net's ability to capture long-range dependencies~\cite{pan2022male,guo2024transpv,xiao2024multiscalegenerativemodelingfast}. In scenes with multiple interacting objects or overlapping structures, this can lead to misclassification or merging of distinct regions~\cite{wu2009detection,nan2012search,rosman2011learning}. Several extensions have been proposed to address these limitations. For instance, Attention U-Net~\cite{oktay2018attention} introduces attention gates to refine the skip connections, allowing the model to selectively emphasize relevant features. Similarly, Residual U-Net~\cite{zhang2018road} incorporates residual connections to facilitate the training of deeper networks. While these modifications improve gradient flow and local feature extraction, they do not fully resolve the challenge of aggregating global context across the entire image.

\subsection{Swin Transformers}
\vspace{-2mm}
Transformer-based models have emerged as powerful alternatives for image segmentation due to their ability to model long-range dependencies through self-attention~\cite{shamshad2023transformers,khan2022transformers,yang2021focal}. Swin Transformers~\cite{liu2021swin} represent a significant development in this area by adopting a hierarchical architecture. The model partitions the input image into non-overlapping patches and computes self-attention within local windows. A key innovation is the introduction of shifted windows between successive layers, which enables cross-window interactions and effectively extends the receptive field without incurring the high computational cost of full global attention. This hierarchical design facilitates multi-scale feature learning, making Swin capable of processing high-resolution images while balancing local detail and global context.

However, these benefits come at a substantial computational cost. The increased memory requirements and processing demands, especially in deeper configurations or with larger input sizes, can be prohibitive for real-time inference or deployment on resource-constrained hardware~\cite{stahl2021deeperthings,jia2020ncpu,mazumder2021survey}. Thus, the trade-off between segmentation performance and efficiency remains an active research challenge for transformer-based methods.

\subsection{Mask2Former}
\vspace{-2mm}
Mask2Former~\cite{cheng2022masked} builds upon the framework established by MaskFormer~\cite{cheng2021per} by reformulating segmentation as a set prediction problem. In MaskFormer, segmentation is achieved by assigning a unique mask to each object instance, thereby unifying the treatment of both “things” and “stuff.” Mask2Former refines this approach by introducing dynamic attention masks that adaptively focus on relevant image regions. This mechanism enables the model to effectively separate overlapping objects and generate high-quality instance and panoptic segmentation outputs.

The dynamic mask generation in Mask2Former facilitates the capture of global contextual information while maintaining the flexibility to delineate object boundaries. However, reliance on attention mechanisms alone may result in the loss of fine-grained local details, which are crucial for accurate boundary delineation, particularly in scenes with small objects or complex textures~\cite{ankareddy2025dense,chu2024fine,zhao2024adaptive}. Despite these challenges, Mask2Former has demonstrated robust performance on standard segmentation benchmarks. Its design underscores the potential of mask-based attention in bridging the gap between local precision and global context, even though the computational complexity and training data requirements continue to be areas for further improvement.

\section{Methods}
\label{sec:method}
\vspace{-2mm}
In this section, we describe our proposed segmentation method in detail. Our approach processes an input image to produce a pixel-wise classification mask, where each pixel is assigned a semantic label. To extend the method’s capabilities, we also incorporate instance and panoptic segmentation branches that leverage shared feature representations and specialized loss functions. We first outline the overall architecture of the model and then explain the training objective and optimization procedure. 

\subsection{Architecture Overview}
\vspace{-2mm}
The MaskAttn-UNet network follows an encoder and a decoder, with mask attention modules integrated at multiple scales. The encoder extracts hierarchical features through successive convolutional blocks that progressively reduce the spatial resolution. At each scale, the features are refined by a mask attention module that generates a learnable binary mask to suppress uninformative regions and emphasize salient structures. Skip connections link corresponding encoder and decoder layers, which helps the decoder recover high-resolution details in the segmentation output. The decoder gradually upsamples and fuses features (augmented by the skip connections) to produce the final prediction.

Fig.~\ref{fig:overview} provides an overview of the architecture. In Fig.~\ref{fig:overview}(a), the overall U-Net style processing pipeline is depicted. Fig.~\ref{fig:overview}(b) illustrates the internal structure of the mask attention module, which applies learnable attention masks to enhance feature representations. Fig.~\ref{fig:overview}(c) shows the detailed multi-scale encoder–decoder design, including the arrangement of convolutional layers, skip connections, and mask attention blocks at each level of resolution.

\subsection{Mask Attention Module}
\label{subsec:maskattn}
\vspace{-2mm}
Each mask attention module is inspired by multi-head self-attention, with an additional learnable mask that modulates the attention weights. Given an input feature map $X$ from either the encoder or decoder, we first reshape it to $X' \in \mathbb{R}^{B\times H\times W\times C}$, where $B$ is the batch size, $H\times W$ are the spatial dimensions, and $C$ is the number of channels. We then apply multi-head masked self-attention (using four heads in our implementation). The attention weights are computed using the scaled dot-product attention mechanism with an added mask matrix $M$:

\begin{equation}
    \centering
    \text{MaskAttn}(Q,K,V,M) = \text{Softmax}\Bigl(\frac{QK^T}{\sqrt{d_k}} + M\Bigr)V
    \label{eq:mask_atten}
\end{equation}
where $Q = X'W^Q$, $K = X'W^K$, $V = X'W^V$, and $d_k$ is the dimensionality of the query and key vectors. Here, $M$ is a learnable (or dynamically computed) mask that suppresses contributions from uninformative regions in the attention matrix. Intuitively, $M$ biases the attention to focus on relevant spatial locations.

The output of the attention operation for a given head is then combined across all heads (as in multi-head attention) and added to the original input via a residual connection. Let $A$ denote the result of the masked multi-head attention (after merging heads). We feed $A$ through a two-layer feed-forward network (FFN) with a GELU nonlinearity, and add the residual $A$ at the end:
\begin{equation}
    \centering
    A_{\text{out}} = \text{GELU}(A W_1 + b_1)W_2 + b_2 + A
    \label{eq:mask_ffn}
\end{equation}
where $W_1, W_2$ are weight matrices and $b_1, b_2$ are biases of the FFN. This yields the final output $A_{\text{out}}$ of the mask attention module. The combination of masked self-attention and the residual FFN enhances the feature representation by integrating global context, while preserving the original information passed through the skip connection.

\subsection{Segmentation Loss}
\label{subsection:loss}
\vspace{-2mm}
We optimize the network using a composite loss function that combines a semantic segmentation loss and an instance-level contrastive loss. Balancing these objectives allows the model to learn both pixel-level class distinctions and instance-specific separability.

\textbf{Semantic Segmentation Loss.}  
For semantic segmentation, where each pixel belongs to one of $C$ classes, we use the standard cross-entropy loss. Let $y_{ij}$ be the ground-truth class label for pixel $(i,j)$, and let $p_{ij}(c)$ be the predicted probability that pixel $(i,j)$ is of class $c$. The loss is:

\begin{equation}
L_{\mathrm{CE}} = -\sum_{i,j} \sum_{c=1}^{C} \delta[y_{ij}=c] \log\bigl(p_{ij}(c)\bigr),
\label{eq:Lce}
\end{equation}

where $\delta[\cdot]$ is the Kronecker delta (which is 1 when its argument is true, and 0 otherwise). This per-pixel cross-entropy encourages correct class predictions for each pixel.

\textbf{Instance Contrastive Loss.}  
For instance segmentation (and the instance component of panoptic segmentation), we employ a contrastive embedding loss that encourages pixels of the same object instance to have similar feature embeddings, while pushing apart embeddings of pixels from different instances. Let $e_{ij}$ denote the embedding vector for pixel $(i,j)$ produced by the network. For a given pixel $(i,j)$, define $P_{ij}$ as the set of positive pixel indices (those belonging to the same ground-truth instance as $(i,j)$), and $N_{ij}$ as the set of negative pixel indices (those belonging to different instances). We first compute a normalizer $D_{ij}$ over all considered pairs for $(i,j)$:

\begin{equation}
D_{ij} = \sum_{(m,n)\in\mathcal{P}_{ij}\cup\mathcal{N}_{ij}} \exp\Bigl(\frac{e_{ij}\cdot e_{mn}}{\tau}\Bigr),
\end{equation}

where $\tau$ is a temperature parameter controlling the sharpness of the contrastive distribution. For a positive pair $(i,j)$ and $(k,l) \in P_{ij}$ (i.e., two pixels from the same instance), the per-pair contrastive loss is:

\begin{equation}
l_{ij,kl} = -\log\frac{\exp\Bigl(\frac{e_{ij}\cdot e_{kl}}{\tau}\Bigr)}{D_{ij}}.
\end{equation}

which penalizes the model if embeddings $e_{ij}$ and $e_{kl}$ are not significantly closer to each other (numerator) compared to all other pairs (denominator). The instance contrastive loss for pixel $(i,j)$ is computed by averaging $l_{ij,kl}$ over all its positive partners $(k,l)\in P_{ij}$, and then averaging over all pixels:

\begin{equation}
L_{\mathrm{IC}} = \frac{1}{N}\sum_{(i,j)} \frac{1}{|\mathcal{P}_{ij}|}\sum_{(k,l)\in\mathcal{P}_{ij}} l_{ij,kl}\,,
\label{eq:Lic}
\end{equation}

where $N$ is the total number of pixels considered (for efficiency, this can be a sampled subset of all pixel pairs). In practice, $L_{\text{IC}}$ encourages embeddings from the same instance to cluster together in feature space, while different-instance embeddings remain separated.The final segmentation loss is a weighted sum of the two components:

\begin{equation}
L_{\mathrm{seg}} = L_{\mathrm{CE}} + \lambda\, L_{\mathrm{IC}},
\label{eq:Lcomb}
\end{equation}

where $\lambda$ controls the balance between the semantic segmentation loss and the instance contrastive loss. In our experiments, we tune $\lambda$ to ensure neither term dominates, enabling the model to learn both accurate pixel-wise classifications and well-separated instance embeddings. (Details on selecting $\lambda$ are provided in Appendix~\ref{sec:lambda}.)

\section{Experiments}
\label{sec:experiments}

\begin{figure*}
    \centering
    \includegraphics[width=\linewidth]{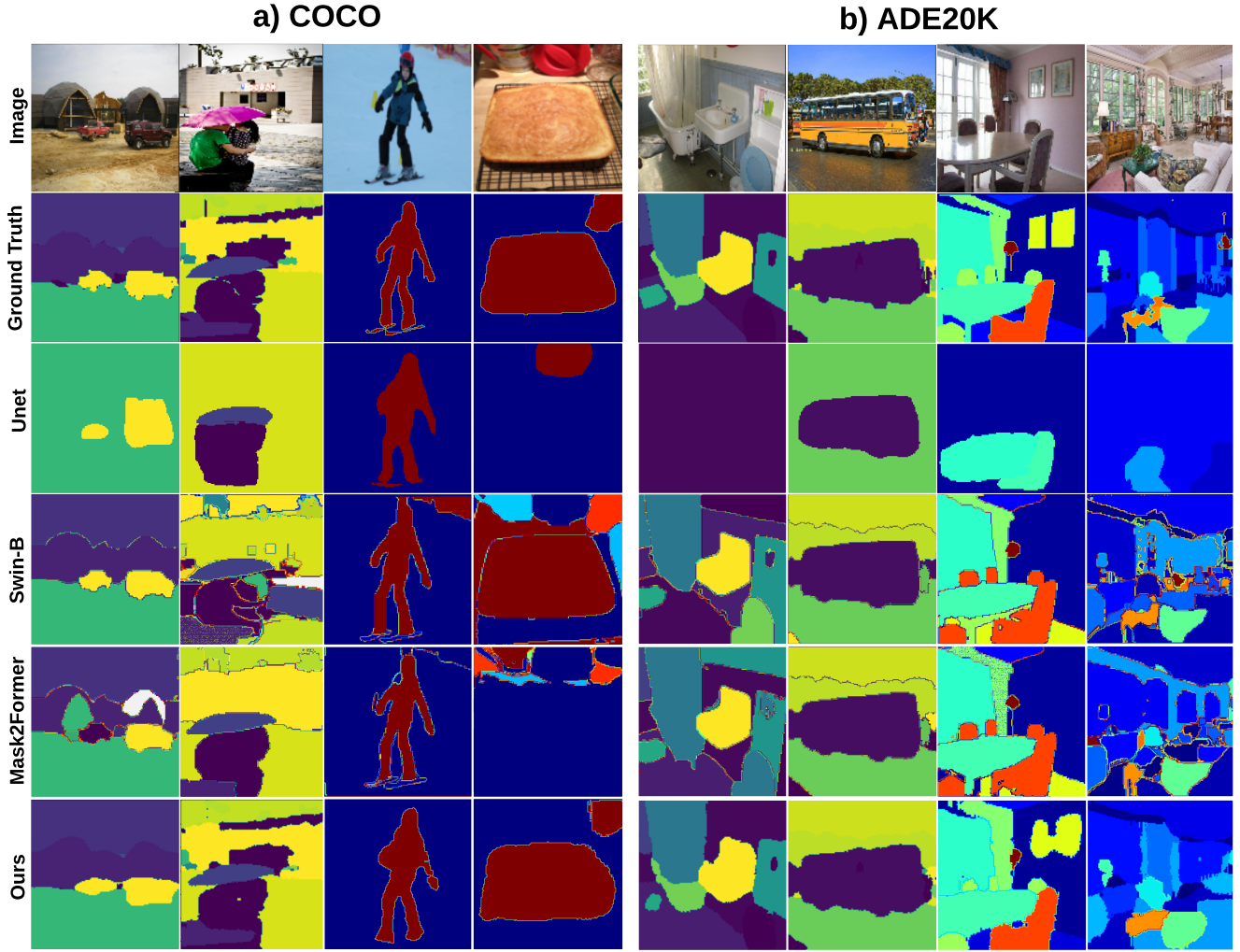}
    \caption{\textbf{Visualization of segmentation results on (a) COCO and (b) ADE20K.} For each dataset, the left two columns show semantic segmentation, and the right two columns show instance segmentation. The top row in each block is the input image, followed by the ground truth, and then predictions from different methods.}
    \label{fig:segmentation}
    \vspace{-5mm}
\end{figure*}

We evaluate MaskAttn-UNet on three commonly used segmentation benchmarks, reporting its semantic, instance, and panoptic segmentation results. Then, we compare our model with state-of-the-art methods on panoptic segmentation and examine its performance using different fractions of the training dataset. These experiments support our design choices for the mask attention modules and demonstrate MaskAttn-UNet’s ability to generalize across datasets.

\textbf{Datasets.}
We employ three widely used benchmarks for multi-task image segmentation. COCO~\cite{lin2014microsoft} is a large-scale dataset with 80 “thing” object categories and multiple “stuff” (background) categories, supporting semantic, instance, and panoptic segmentation tasks. ADE20K~\cite{zhou2017scene} includes 150 semantic categories (100 “things” and 50 “stuff”) and is used for semantic, instance, and panoptic segmentation. Cityscapes~\cite{cordts2016cityscapes} focuses on urban street scenes with 19 classes (8 “thing” and 11 “stuff”), commonly used for semantic and panoptic segmentation in autonomous driving scenarios. All images are resized to $128\times128$ pixels to reduce computational overhead and simulate low-resolution conditions encountered in certain real-world applications.

\textbf{Evaluation Metrics.}
We use standard metrics for each task. 
For \textit{Semantic Segmentation}, we report mean Intersection-over-Union (mIoU)~\cite{everingham2015pascal}, which measures the average per-class overlap between predicted and ground-truth regions.
For \textit{Instance Segmentation}, we use the Average Precision (\textbf{AP})~\cite{lin2014microsoft} at various intersection-over-union thresholds, which evaluates how well individual object instances are detected and segmented (higher AP indicates better precision-recall tradeoff).
For \textit{Panoptic Segmentation}, we report the Panoptic Quality (\textbf{PQ}) metric~\cite{kirillov2019panoptic}, which encapsulates both recognition quality (RQ) and segmentation quality (SQ) for the combined set of “thing” and “stuff” classes. Along with mIoU and AP under 100\% mIoU threshold. These metrics provide a comprehensive evaluation of segmentation performance on each dataset.

\subsection{Implementation Details}
\label{subsec:implementation}

Our implementation builds on a U-Net backbone with four downsampling encoder stages and four upsampling decoder stages, connected by skip connections to recover spatial detail. The encoder gradually increases the number of feature channels from 64 to 128 to 256 (with two blocks at 256), leading into a bottleneck that compresses and refines the global context. The decoder then symmetrically upsamples, reducing the channel dimensions and merging feature maps from the corresponding encoder levels to reconstruct fine-grained spatial details in the output.

At every encoder and decoder stage, we incorporate a MaskAttn module as described in Section~\ref{subsec:maskattn}. Each module contains a learnable binary mask that selectively suppresses non-relevant activations. This directs the network’s attention to important regions, such as object boundaries and salient structures, even in low-resolution feature maps. By integrating these modules throughout the network, MaskAttn-UNet retains the inductive bias of convolutions for locality while gaining the ability to capture long-range dependencies at each scale (More detailed analysis can be found in Appendix~\ref{sec:long_range}).

\begin{table}[ht]
\vspace{-2mm}
\centering
\begin{tabular}{lccc}
\toprule
\textbf{Metric} & \textbf{COCO} & \textbf{ADE20K} & \textbf{Cityscapes} \\
\midrule
mIoU$\uparrow$ (\%)  & 43.7 & 44.1 & 67.4 \\
\bottomrule
\end{tabular}
\caption{\textbf{Semantic segmentation on COCO (133 categories), ADE20K (150 categories), and Cityscapes (19 categories).} Our model consistently achieved remarkable results on all kinds of data.}
\label{tab:semantic_results}
\vspace{-6mm}
\end{table}

\subsection{Training Settings}

\textbf{Loss Functions.}
We train all branches of the model (semantic, instance, panoptic) separately by choosing the losses defined in Section~\ref{subsection:loss}. For semantic segmentation, we apply the cross-entropy loss (Eq.~\ref{eq:Lce}), which provides a strong per-pixel classification signal. For instance and panoptic segmentation, we include the instance contrastive loss (Eq.~\ref{eq:Lic}) to learn distinct object embeddings. In practice, we also add standard classification loss terms for each predicted instance (to predict its semantic category), ensuring that each instance embedding is associated with a specific class. By balancing the contributions of $L_{\text{CE}}$ and $L_{\text{IC}}$ ($\lambda=0.5$ was chosen in Eq.~\ref{eq:Lcomb}), the model learns both class-level distinctions and fine-grained instance separation simultaneously.

\textbf{Training Setup.}
 We train our models for 1000 epochs on each dataset. All images are uniformly resized to $128\times128$. For semantic segmentation experiments, we use $2\times$30GB NVIDIA V100 GPUs with a total batch size of 8. For the more memory-intensive instance and panoptic segmentation experiments, we use $2\times$40GB NVIDIA A100 GPUs with a batch size of 14. The model is optimized using the AdamW optimizer with an initial learning rate of $5\times10^{-4}$ and a weight decay of $10^{-3}$. We employ data augmentation techniques, including random scale jittering and horizontal flipping, to provide input diversity without excessively complicating the training distribution. These training configurations are chosen to balance memory usage and throughput for each task, resulting in consistent convergence across the different segmentation objectives.

\begin{table}[ht]
\centering
\begin{tabular}{lcccc}
\toprule
\textbf{Dataset} & \textbf{AP@30} & \textbf{AP@50} & \textbf{AP@70} & \textbf{AP@100}\\
\midrule
COCO       & 35.0 & 31.6 & 31.3 & 30.2 \\
ADE20K     & 33.8 & 33.2 & 30.5 & 30.5 \\
Cityscapes & 38.9 & 36.6 & 36.2 & 35.5 \\
\bottomrule
\end{tabular}
\caption{\textbf{Instance segmentation results (AP@k).} k stands for mIoU threshold. MaskAttn-Unet kept stable even with a small threshold. This indicates that the proposed model reliably isolates and delineates individual object instances.}
\label{tab:instance_results}
\end{table}
\vspace{-5mm}

\subsection{Main Results}
\label{subsec:main_results}


\textbf{Semantic Segmentation.}  
We evaluate our semantic segmentation model using the COCO \texttt{panoptic\_val2017}, ADE20K \texttt{val}, and Cityscapes \texttt{val} datasets, with all labels processed specifically for semantic segmentation. As shown in Tab.~\ref{tab:semantic_results}, MaskAttn-UNet achieves a mIoU of 43.7\% on COCO, 44.1\% on ADE20K, and 67.4\% on Cityscapes. These results demonstrate that the network efficiently fuses global contextual information with local spatial details despite the low resolution of the inputs. In the left two columns of Fig.~\ref{fig:segmentation}(a) and Fig.~\ref{fig:segmentation}(b), the segmentation maps reveal that object boundaries are well preserved and regions with complex textures are segmented with clarity. In particular, areas containing overlapping objects or fine structural details are handled effectively. This suggests that the mask attention mechanism successfully suppresses background noise and enhances the discriminative power of learned features.

\begin{table}[ht]
\centering
\begin{tabular}{lccc}
\toprule
\textbf{Dataset} & \textbf{mIoU$\uparrow$ (\%)} & \textbf{AP$\uparrow$ (\%)} & \textbf{PQ$\uparrow$ (\%)} \\
\midrule
COCO       & 45.3 & 31.5 & 35.7 \\
ADE20K     & 45.9 & 30.7 & 33.6 \\
Cityscapes & 70.1 & 35.5 & 58.3 \\
\bottomrule
\end{tabular}
\caption{\textbf{Panoptic segmentation performance on COCO, ADE20K, and Cityscapes.} MaskAttn-UNet demonstrates robust segmentation capabilities across diverse datasets.}
\label{tab:panoptic_results}
\vspace{-5mm}
\end{table}

\begin{table*}[h]
\centering
\begin{adjustbox}{max width=\textwidth}
\begin{tabular}{lccccccccccc}
\toprule
\multirow{2}{*}{\textbf{Method}} & \multirow{2}{*}{\textbf{FLOPs (G)}} & \multirow{2}{*}{\textbf{\#Params (M)}} & 
\multicolumn{3}{c}{\textbf{COCO}} & \multicolumn{3}{c}{\textbf{ADE20K}} & \multicolumn{3}{c}{\textbf{Cityscapes}} \\ 
\cmidrule(lr){4-6} \cmidrule(lr){7-9} \cmidrule(lr){10-12}
& & & mIoU$\uparrow$ & PQ$\uparrow$ & AP$\uparrow$ & mIoU$\uparrow$ & PQ$\uparrow$ & AP$\uparrow$ & mIoU$\uparrow$ & PQ$\uparrow$ & AP$\uparrow$ \\
\midrule
U-Net       & 4  & 32  & 33.8 & 20.1 & 13.8 &37.3 & 24.1 & 16.5 & 63.4 &51.2 & 27.9 \\
DETR-R50 & 86 & 41 & 34.7 & 25.3 & 16.7 & 40.7 & 26.5 & 21.0 & 66.6 & 53.7 & 30.6\\
DETR-R101 & 152 & 60 & 35.3 & 27.4 & 18.8 & 41.1 & 27.4 & 21.9 & 67.4 &54.2 & 32.2\\
Mask2Former-R50 & 226  & \textbf{44}  & 38.1 & 31.4 & 27.3 & 45.2 & 33.4 & 29.1 & 69.3 & 55.7 & 34.3 \\
Mask2Former-R101&293 &63 & \underline{39.2} & \underline{31.7} & \underline{28.1} & \underline{45.7} & \textbf{33.9} & \underline{30.5} & \textbf{71.4}& \underline{57.2} & \underline{35.4}\\
\midrule
\textbf{MaskAttn-UNet}(Ours)  & \textbf{11} & 46  & \textbf{45.3} & \textbf{35.7} & \textbf{31.5} & \textbf{45.9} & \underline{33.6} & \textbf{30.7} & \underline{70.1} & \textbf{58.3} & \textbf{35.5} \\
\bottomrule
\end{tabular}
\end{adjustbox}
\caption{\textbf{Comparison of \textbf{MaskAttn-UNet} with state-of-the-art models for panoptic segmentation} MaskAttn-UNet achieves competitive segmentation performance with significantly lower computational complexity, as indicated by its reduced FLOPs and parameter count. This efficiency highlights its potential for applications requiring a balance between accuracy and resource constraints. The best results are highlighted in \textbf{bold}, and the second best are \underline{underlined}.}
\label{tab:comparison_baselines}
\vspace{-5mm}
\end{table*}

Furthermore, a closer inspection of the segmentation results shows that the network adapts well to varying scene complexities. In particular, in images with diverse lighting and contrast conditions (First image in Fig.~\ref{fig:segmentation}(b)), the network consistently maintains high accuracy, ensuring that both large homogeneous regions and small, intricate details are accurately labeled. The design of the mask attention modules appears to selectively amplify important features while reducing interference from less informative areas, thereby yielding more consistent predictions across different classes and challenging scenarios.



\textbf{Instance Segmentation.}
Instance segmentation performance was assessed on the COCO \texttt{val2017}, ADE20K \texttt{val}, and Cityscapes \texttt{val} datasets, with all labels refined to suit the task. Table~\ref{tab:instance_results} summarizes the performance of MaskAttn-UNet. On COCO, the model achieves an Average Precision (AP) of 35.0\% at an IoU threshold of 30, decreasing to 30.2\% at an IoU threshold of 100. ADE20K yields AP values of 33.8\%, 33.2\%, 30.5\%, and 30.5\% for IoU thresholds of 30, 50, 70, and 100, respectively, while Cityscapes reports corresponding AP values of 38.9\%, 36.6\%, 36.2\%, and 35.5\%. These results indicate that MaskAttn-UNet reliably isolates and delineates individual object instances, even in scenarios with overlapping or densely arranged objects.

The effect of varying IoU thresholds provides insight into the network's ability to handle different instance complexities. Lower IoU thresholds primarily capture the most prominent objects, resulting in higher AP values, whereas higher IoU thresholds extend detection to smaller or partially occluded instances, leading to a slight reduction in AP. Visualizations of the instance segmentation outputs, shown in the right two columns of Fig.~\ref{fig:segmentation}(a) and Fig.~\ref{fig:segmentation}(b), indicate that the mask attention modules help refine feature maps at multiple scales, improving the separation of adjacent objects and the recognition of fine structural details. For example, the segmentation result for the third image in Fig.~\ref{fig:segmentation}(b) shows that MaskAttn-UNet reduces misdetections in areas with overlapping objects and intricate boundaries. The network preserves object contours in cluttered regions, maintaining consistent separation of individual instances. These results demonstrate that the proposed architecture adapts well to various real-world conditions, reinforcing its effectiveness for practical instance segmentation applications.

\textbf{Panoptic Segmentation.} Panoptic segmentation performance was evaluated on the COCO \texttt{panoptic\_val2017}, ADE20K \texttt{val}, and Cityscapes \texttt{val} datasets. Table~\ref{tab:panoptic_results} presents the results, where MaskAttn-UNet achieves a mean Intersection-over-Union (mIoU) of 45.3\% for "stuff" regions, an Average Precision (AP) of 31.5\% for "thing" instances, and an overall Panoptic Quality (PQ) of 35.7\% on COCO. On ADE20K, the model attains a mIoU of 45.9\%, an AP of 30.7\%, and a PQ of 33.6\%. Similarly, on Cityscapes, it records values of 70.1\% for mIoU, 35.5\% for AP, and 58.3\% for PQ. These outcomes indicate that MaskAttn-UNet delivers balanced segmentation performance across both foreground objects and background regions.

The high mIoU values on Cityscapes suggest that the network effectively leverages the structured nature of urban scenes to achieve consistent background segmentation. Meanwhile, the stable AP and PQ values on COCO and ADE20K demonstrate its ability to handle more diverse and complex environments. The integration of robust background segmentation with precise instance delineation contributes to a high overall PQ, affirming the network's capability to provide comprehensive scene understanding.

\textbf{Comparison with Baseline Models.}
To validate the robustness of our approach, we benchmarked our model against several state-of-the-art models with comparable or slightly greater complexity across three datasets: COCO, ADE20K, and Cityscapes (see Tab.~\ref{tab:comparison_baselines}). For instance, compared to U-Net, which operates at $4G$ FLOPs with $32M$ parameters, MaskAttn-UNet delivers improvements of over \textbf{10\%} in mIoU, \textbf{15\%} in PQ, and nearly \textbf{20\%} in AP on COCO, while only increasing the parameter count by about $15M$. Although U-Net’s lower FLOPs are beneficial for simpler scenes, it struggles in environments with overlapping objects and complex textures. In contrast, MaskAttn-Unet, with $11G$ FLOPs and $46M$ parameters, effectively captures fine spatial details and manages cluttered scenes more efficiently.

We also evaluated DETR-based models, including DETR-R50 ($86G$ FLOPs, $41M$ parameters) and DETR-R101 ($152G$ FLOPs, $60M$ parameters).  DETR-based models similarly require increased computing with only moderate performance gains, limiting their applicability in resource-constrained or real-time settings. 

Moreover, Mask2Former-R50 demands a substantial  $226G$ FLOPs, while Mask2Former-R101, employing a ResNet-101 backbone, requires even more computational resources. Notably, Mask2Former-R101 achieves only marginal improvements over MaskAttn-UNet—it records a 0.3\% higher PQ on ADE20K and a 1.3\% higher mIoU on Cityscapes. However, these slight gains come at a steep cost: Mask2Former-R101 uses $17M$ more parameters than MaskAttn-UNet (roughly a 37\% increase), highlighting that simply increasing compute does not necessarily translate to substantially better segmentation quality. Overall, MaskAttn-UNet achieves a strong balance between accuracy and efficiency, making it a practical choice for applications with constrained computational resources.

\begin{figure}[!htbp]
    \centering
 \includegraphics[width=\linewidth]{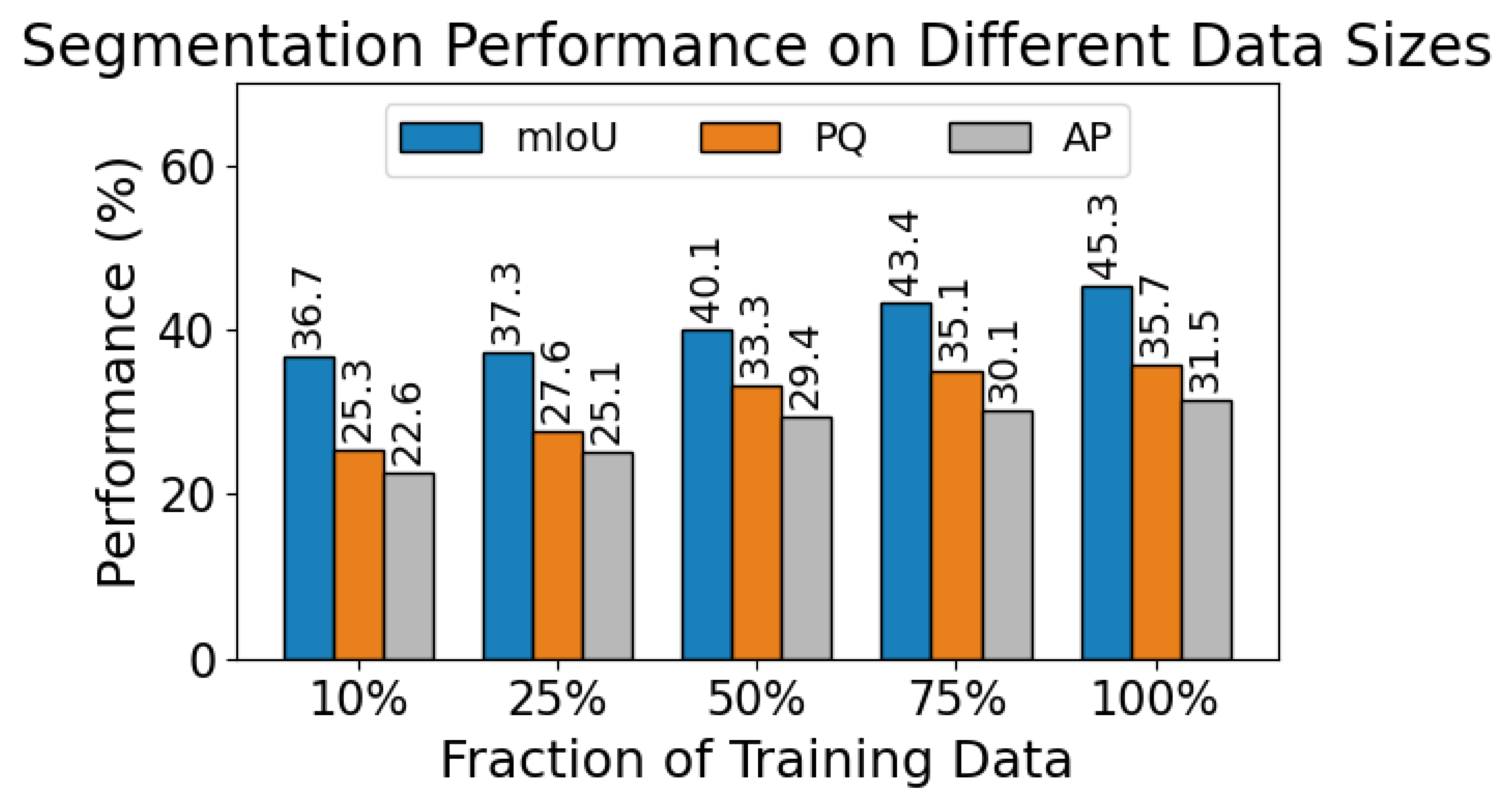}
    \caption{\textbf{Segmentation performance of MaskAttn-UNet on different fractions (10\%, 25\%, 50\%, 75\%, 100\%) of the \texttt{panoptic\_train2017} dataset.} Results illustrate consistent improvement across metrics with increasing dataset size, highlighting the model's strong data efficiency.}

    \label{fig:frac_data}
     \vspace{-2mm}
\end{figure}

\raggedbottom
\textbf{Few Shots Training.} In many real-world settings, collecting extensive annotated datasets is both expensive and time-consuming. To assess the effect of training data size on segmentation performance, we evaluated MaskAttn-UNet on the COCO \texttt{panoptic\_val2017} dataset, training the model with varying fractions (10\%, 25\%, 50\%, 75\%, and 100\%) of the full dataset. The results, depicted in Fig.~\ref{fig:frac_data}, reveal the following trends: 
\begin{itemize}
    \item \textbf{10\% Training Data:} The model achieves 36.7\% mIoU, 25.3\% PQ, and 22.6\% AP. While these metrics are suboptimal, they demonstrate the network's capacity to extract meaningful features even under data-scarce conditions.
    \item \textbf{25\% Training Data:} Performance improves to 37.3\% mIoU, 27.6\% PQ, and 25.1\% AP, indicating that a modest increase in training data leads to significant gains in segmentation accuracy.
    \item \textbf{50\% Training Data:} The model attains 40.1\% mIoU, 33.3\% PQ, and 29.4\% AP, suggesting that half of the full dataset suffices for learning robust feature representations.
    \item \textbf{75\% Training Data:} Metrics further rise to 43.4\% mIoU, 35.1\% PQ, and 30.1\% AP, confirming that additional data continues to benefit the model while maintaining efficient learning dynamics.
    \item \textbf{100\% Training Data:} Utilizing the entire dataset, the model achieves 45.3\% mIoU, 35.7\% PQ, and 31.5\% AP, highlighting its capacity to fully exploit available annotations for optimal segmentation results.
\end{itemize}

These trends underline the strong data efficiency of MaskAttn-UNet, making it a practical choice for circumstances with limited annotated data. Notably, the model exhibits substantial performance improvements with increasing data availability, aligning with established neural scaling laws that describe how neural network performance scales with dataset size ~\cite{alabdulmohsin2022revisiting,kaplan2020scaling}. Such data efficiency is particularly invaluable in medical image computing and other fields where acquiring large, annotated datasets is challenging \cite{simpson2019large,razzak2017deep}. Therefore, MaskAttn-UNet's ability to perform effectively with reduced training data positions it as a viable solution in data-constrained environments.

\raggedbottom
\vspace{-2mm}
\section{Discussion}
\label{sec:discussion}
\vspace{-2mm}

We present MaskAttn-UNet, a significant advancement in segmentation models by integrating masked attention modules into the traditional U-Net architecture, which effectively enhances both local and global feature extraction. This hybrid approach leverages the strengths of convolutional networks in modeling local context and the capabilities of masked attention mechanisms for long-range dependencies. Our empirical evaluations on datasets such as COCO, ADE20K, and Cityscapes demonstrate that MaskAttn-UNet consistently outperforms standard U-Net models while utilizing significantly fewer computational resources compared to transformer-based architectures like Mask2former. These findings highlight the potential of selective attention mechanisms in low-resolution segmentation tasks, bridging the gap between convolutional efficiency and the global context awareness characteristic of transformer models.

Despite these promising results, there are avenues for further improvement. Future work will focus on extending MaskAttn-UNet to domains such as medical imaging, where data scarcity and the need for precise boundary delineation present unique challenges. Additionally, integrating MaskAttn-UNet into diffusion-based models could enhance generative and data augmentation processes. Addressing the segmentation of small objects and intricate details remains an active area of research; incorporating specialized modules or tailored loss functions could further elevate performance. These future directions aim to refine MaskAttn-UNet's capabilities and broaden its applicability across diverse real-world segmentation scenarios.
{
    \small
    \bibliographystyle{ieeenat_fullname}
    \bibliography{main}
}

\clearpage
\appendix

\section*{Appendix}
\addcontentsline{toc}{section}{Appendix}

\label{sec:supp}
\setcounter{section}{0}

\section{Code and Data Availability}

Our code to run the experiments can be found at  \url{https://anonymous.4open.science/r/MaskUnet-736B}.

\section{More Experiments and Dicussion}
\label{sec:more}

\subsection{Effect of $\lambda$ on the Loss Function}
\label{sec:lambda}

\begin{figure}[H]
    \centering
    \includegraphics[width=0.8\linewidth]{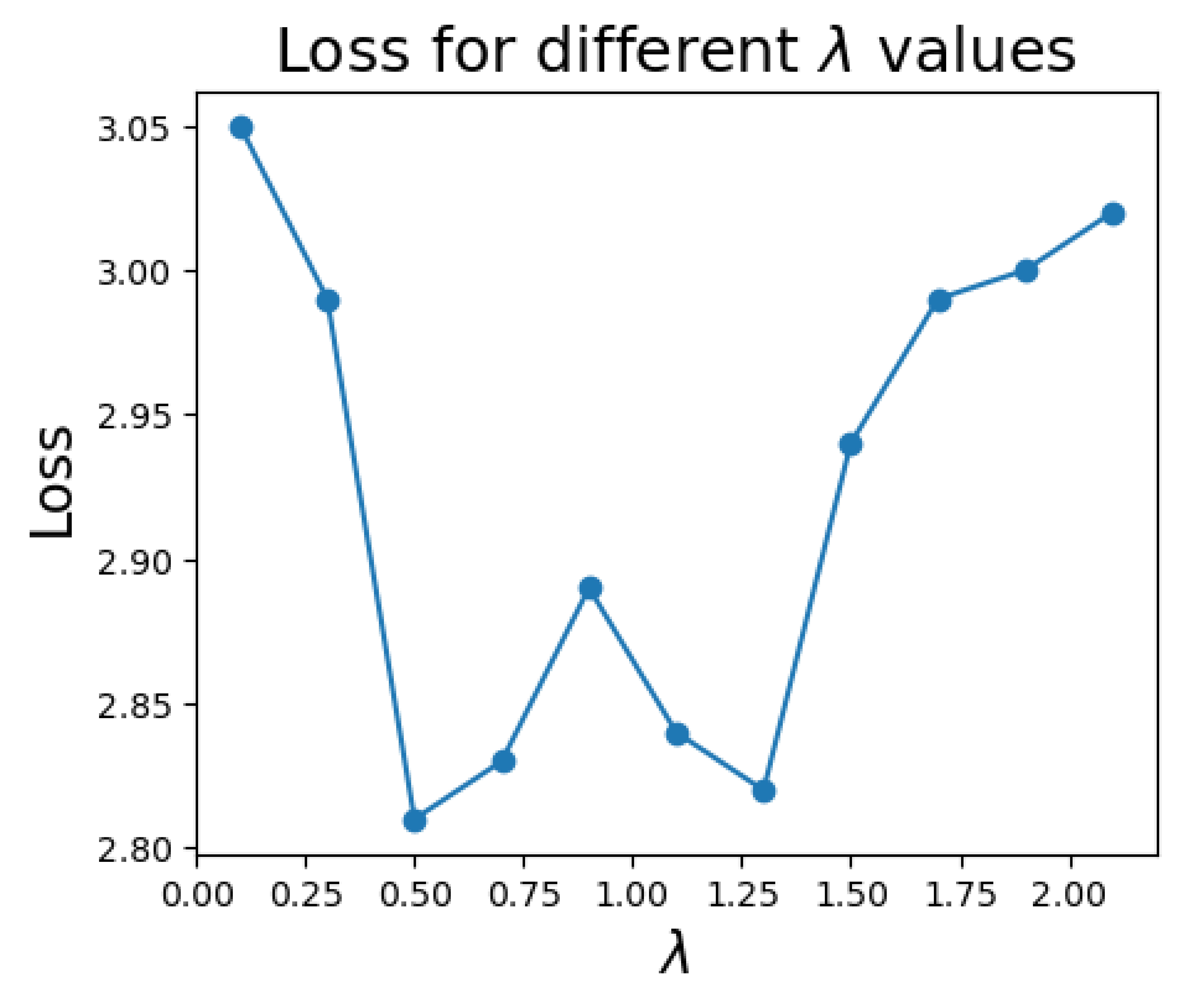}
    \caption{\textbf{Trend of the combined loss as a function of $\lambda$.} The model only ran 20 epochs, not fully trained. The minimum loss is observed at $\lambda=0.5$.}
    \label{fig:lambda}
\end{figure}

As indicated in Eq.~\ref{eq:Lcomb}, the parameter $\lambda$ governs the relative influence of the cross-entropy loss and the instance contrastive loss. We explored a range of values from 0.1 to 2.1, training MaskAttn-UNet for 20 epochs to observe how different $\lambda$ settings affect the combined loss. Fig.~\ref{fig:lambda} illustrates that the loss consistently decreases and reaches its lowest point at $\lambda=0.5$. When $\lambda$ is too small (e.g., below 0.3), the instance contrastive term is underemphasized, leading to weaker instance separation. Conversely, larger values of $\lambda$ (above 1.0) can overshadow the cross-entropy component, resulting in suboptimal pixel-level classifications.

These findings indicate that setting $\lambda=0.5$ provides a well-balanced combination of the cross-entropy and instance contrastive losses for instance and panoptic segmentation. In this configuration, MaskAttn-UNet achieves reliable pixel-level predictions and effective instance discrimination, which are critical for accurately delineating overlapping objects and complex scene structures.

Although the model was trained for only 20 epochs, the early-stage trends in the combined loss clearly show that this limited training is sufficient to assess the impact of $\lambda$. The minimum loss observed at $\lambda=0.5$ suggests that this value reasonably balances the two loss components, providing a solid basis for selecting it as the optimal setting. Further training and a finer-grained parameter search are expected to fine-tune rather than dramatically alter this optimal balance.

\subsection{Ablation Studies}
\paragraph{MaskAttn Module.}
To assess the contribution of Mask Attention modules within our MaskAttn-UNet architecture, we conducted an ablation study by systematically removing these modules and evaluating the performance of the resulting baseline UNet. Both models were trained and tested on the COCO \texttt{panoptic\_val2017} dataset, and their performance metrics are detailed in Table \ref{tab:ablation_results}.

\begin{table}[h]
\centering
\begin{tabular}{lccc}
\hline
\textbf{Model} & \textbf{mIoU (\%)} & \textbf{PQ (\%)} & \textbf{AP (\%)} \\
\hline
MaskAttn-UNet & 45.3 & 35.7 & 31.5 \\
UNet & 33.8 & 20.1 & 13.8 \\
\hline
\end{tabular}
\caption{Performance comparison between MaskAttn-UNet and baseline UNet on COCO \texttt{panoptic\_val2017} dataset.}
\label{tab:ablation_results}
\end{table}

The integration of Mask Attention modules led to substantial improvements across all evaluated metrics. Specifically, MaskAttn-UNet achieved a mean Intersection over Union (mIoU) of 45.3\%, which is an \textbf{11.5\%} relative increase over the baseline UNet's 33.8\%. In terms of Panoptic Quality (PQ), MaskAttn-UNet reached 35.7\%, marking a \textbf{15.6\%} relative improvement compared to the baseline's 20.1\%. Additionally, the Average Precision (AP) saw a significant enhancement, with MaskAttn-UNet obtaining 31.5\%, corresponding to a \textbf{17.7\%} incline over the baseline's 13.8\%.

The remarkable improvements in mIoU, PQ, and AP suggest that Mask Attention modules enable the network to better capture spatial dependencies and contextual information, which are crucial for accurate segmentation. The significant enhancement in PQ and AP metrics indicates that the MaskAttn-UNet is particularly effective in distinguishing and accurately segmenting individual objects within complex scenes, a task where the baseline UNet exhibits limitations.

\paragraph{Partial MaskAttn Module.}
To evaluate the individual contributions of the attention mechanisms in our MaskAttn-UNet, we conducted experiments with two model variants: one that incorporates Mask Attention modules solely in the encoder and another that applies them only in the decoder. Table~\ref{tab:enc_dec} presents the results on the COCO \texttt{panoptic\_val2017} dataset.

\begin{table}[h]
\centering
\begin{tabular}{lccc}
\hline
\textbf{Variant} & \textbf{mIoU (\%)} & \textbf{PQ (\%)} & \textbf{AP (\%)} \\
\hline
Encoder-Only & 38.1 & 24.3 & 18.8 \\
Decoder-Only & 36.7 & 23.6 & 17.0 \\
\hline
\end{tabular}
\caption{Performance of MaskAttn-UNet variants with attention applied exclusively in the encoder or decoder.}
\label{tab:enc_dec}
\end{table}

Although the encoder-only variant yields slightly better performance compared to the decoder-only variant, both attention modules play complementary roles that are critical for the overall performance of MaskAttn-UNet. The encoder attention modules are particularly effective for low-resolution images, as they enhance the extraction of local details and fine features, which significantly improves segmentation quality. On the other hand, decoder attention modules ensure that long-range dependencies are maintained during the reconstruction process, preserving global contextual information. Both variants surpassed traditional UNet by more than 5\% in mIoU, 3\% larger in PQ, and 5\% incline in AP.

The full MaskAttn-UNet, which integrates attention in both the encoder and decoder, outperforms the individual variants by effectively combining local detail enhancement with global feature preservation. This synergy confirms that both encoder and decoder attention mechanisms are essential for achieving superior segmentation performance, as evidenced by the full model's results (mIoU = 45.3\%, PQ = 35.7\%, AP = 31.5\%).

Overall, our ablation study demonstrates that while encoder attention offers notable benefits on its own, the incorporation of both encoder and decoder attention modules is crucial for capturing the complete spectrum of spatial dependencies, leading to significant improvements in panoptic segmentation.

\paragraph{Mask Layer.}

\begin{table}[h]
\centering
\small
\begin{tabular}{lccc}
\hline
\textbf{Model} & \textbf{mIoU (\%)} & \textbf{PQ (\%)} & \textbf{AP (\%)} \\
\hline
Self-Attention UNet~\cite{petit2021u} & 41.2 & 30.9 & 26.7 \\
MaskAttn-UNet (Ours)             & 45.3 & 35.7 & 31.5 \\
\hline
\end{tabular}
\caption{Comparison of segmentation performance between Self-Attention UNet and MaskAttn-UNet on COCO \texttt{panoptic\_val2017}.}
\label{tab:mask_layer}
\end{table}

We also want to explore how the mask layer improves the attention module in image segmentation tasks. To explore this, we trained the self-attention Unet~\cite{petit2021u} on the same dataset, and the results are shown in Tab.~\ref{tab:mask_layer}. The self-attention Unet achieved a mIoU of 41.2\%, a PQ of 30.9\%, and an AP of 26.7\%. In contrast, our MaskAttn-UNet, which integrates a dedicated mask layer into the attention module, obtained significantly improved metrics. The mask layer acts as a filter that suppresses irrelevant regions and reinforces salient spatial features, thereby addressing common limitations of self-attention in capturing fine-grained local details. This mechanism is particularly effective for low-resolution images, where precise local feature extraction is critical while still preserving long-range dependencies to maintain global context. 

\begin{figure*}
    \centering
    \includegraphics[width=0.8\linewidth]{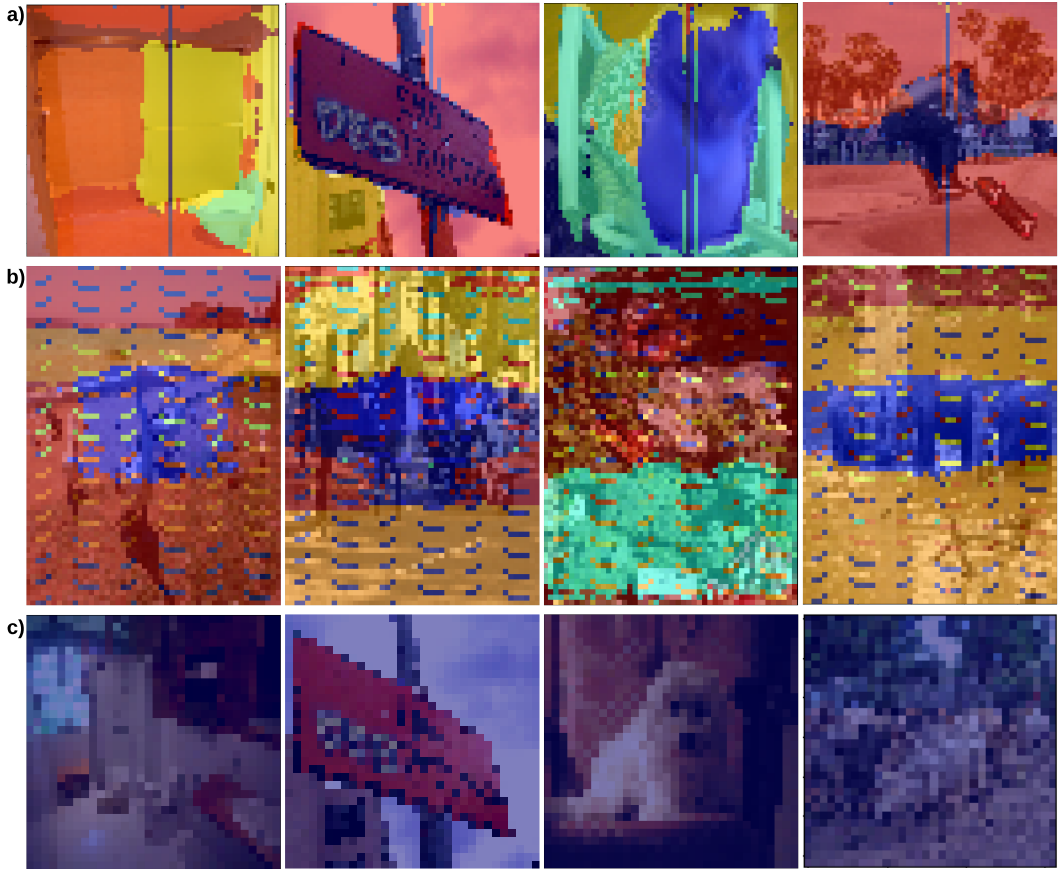}
    \caption{\textbf{Visualization of low-resolution segmentation results}(a) Sample semantic segmentation on $64\times64$ resolution. (b) Semantic segmentation on $64\times48$ resolution.(c) Semantic segmentation on $32\times32$ resolution.}
    \label{fig:low_res}
\end{figure*}

\subsection*{Segmentation on Ultra-Low Image Resolutions}

In many real-world applications, images are captured at resolutions lower than $128\times128$. Consequently, we tested our model at ultra-low resolutions to determine the minimal resolution at which reliable segmentation can be achieved. In practice, many low-cost UAVs and cameras capture images at resolutions of $64\times64$, $64\times48$, and $32\times32$~\cite{reibaldi2010standard,pereira2022sleap}. For baseline comparison, we selected Mask2Former-R50 and Mask2Former-R101 because they demonstrated comparable performance on $128\times128$ segmentation tasks. Table~\ref{tab:low_res} reports the semantic segmentation performance (mIoU) of these models at ultra-low resolutions. Notably, our model is able to segment prominent objects at a resolution of $64\times48$, whereas the Mask2Former models achieve only around 10\% mIoU at this resolution. At the extremely low resolution of $32\times32$, none of the models can produce reliable segmentation results, indicating that this resolution is below the effective threshold. Overall, these results suggest that our model maintains segmentation capabilities at lower resolutions where the baseline methods already struggle, thereby demonstrating its robustness in resource-constrained environments.

\begin{table}[ht]
\small
\centering
\begin{adjustbox}{max width=\textwidth}
\begin{tabular}{lccc}
\toprule
\multirow{2}{*}{\textbf{Method}} & \multicolumn{3}{c}{\textbf{Input Resolution}} \\
\cmidrule(lr){2-4}
& $64\times64$ & $64\times48$ & $32\times32$ \\
\midrule
Mask2Former-R50 & 31.4& 8.7 & 1.3   \\
Mask2Former-R101 & 33.5& 11.2 & 1.9  \\
\textbf{MaskAttn-UNet(Ours)} & 41.3 & 17.1 & 2.2   \\
\bottomrule
\end{tabular}
\end{adjustbox}
\caption{\textbf{Semantic segmentation performance (mIoU) across different input resolutions.} Our model reached resolution limits when resolution less than $64\times48$, whether other models struggled at $64\times64$ resolution.}
\label{tab:low_res}
\end{table}

The visualization in Fig.~\ref{fig:low_res} highlights the capability of MaskAttn-UNet to generate meaningful semantic segmentation results even at extremely low resolutions. At the $64\times64$ resolution (Fig.~\ref{fig:low_res}(a)), object boundaries and primary structures remain clearly distinguishable, illustrating the model's robustness in retaining crucial semantic information despite reduced input size. Remarkably, at the even lower resolution of $64\times48$ (Fig.~\ref{fig:low_res}(b)), MaskAttn-UNet still successfully identifies primary objects, demonstrating its potential to capture coarse semantic cues under highly constrained conditions. 

Although the segmentation at $32\times32$ (Fig.~\ref{fig:low_res}(c)) resolution shows considerable information loss, the model can nonetheless roughly discern dominant objects, confirming its ability to leverage limited pixel information. These qualitative results reinforce the numerical performance trends, supporting the suitability of MaskAttn-UNet for practical applications where computational resources or sensor capabilities are severely limited.

\subsection{Analysis of Long-Range Dependency Capture}
\label{sec:long_range}

In order to quantitatively assess the long-range dependency capture of MaskAttn-UNet, we analyzed feature maps extracted from four key attention modules: \textit{att1} (early encoder), \textit{att3} (bottom encoder), \textit{att4} (bottom decoder), and \textit{att6} (top decoder). For each module, we computed the Hurst exponent,  the scaling exponent from detrended fluctuation analysis (DFA), and the power spectral density (PSD) over a large subset of the COCO \texttt{panoptic\_val2017} dataset.

\textbf{Hurst Exponent.}
The Hurst exponent ($H$) quantifies the presence of long-term memory in time series data~\cite{hurst1951long,mandelbrot1969robustness}. It evaluates whether a series tends to regress to the mean or shows persistent clustering~\cite{beran2010long, carbone2004time}. Formally, the value of $H$ ranges from 0 to 1, with:
\begin{itemize}
\item $H = 0.5$: representing a random walk with no long-range correlation.
\item $H > 0.5$: indicating persistent behavior, meaning that high values tend to follow high values, and low values tend to follow low values.
\item $H < 0.5$: indicating anti-persistent behavior, where values alternate frequently.
\end{itemize}

\begin{figure}[ht]
    \centering
    \includegraphics[width=\linewidth]{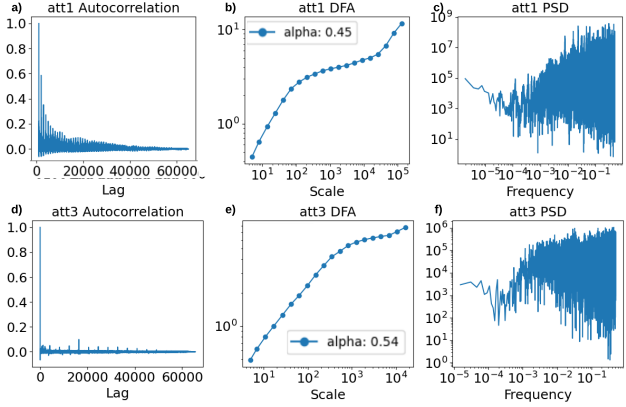}
   \caption{\textbf{Long-Range Dependency Analysis in the Encoder.} (a) Autocorrelation plot for the early encoder layer (\textit{att1}) illustrating the decay of correlation with increasing lag. (b) DFA plot for \textit{att1}, with the slope indicating the scaling behavior. (c) Power spectral density (PSD) plot for \textit{att1} showing the distribution of power across frequencies. (d) Summary visualization of the Hurst exponent across encoder layers. (e) Summary visualization of the DFA exponent across encoder layers. (f) Combined metric overview for encoder long-range dependencies.}
    \label{fig:encoder}
\end{figure}

\textbf{Detrended Fluctuation Analysis (DFA).}
DFA is a robust statistical method for identifying long-range correlations in non-stationary signals by analyzing fluctuations at multiple scales~\cite{peng1994mosaic,bryce2012revisiting,hu2001effect}. The main steps are:
\begin{enumerate}
\item Compute the integrated series by subtracting the mean from the original signal and taking the cumulative sum.
\item Segment this series into equal-length segments.
\item Fit and subtract a polynomial trend (typically linear) from each segment.
\item Calculate the root-mean-square fluctuation of each detrended segment.
\item Analyze the scaling relationship between segment lengths and average fluctuations.
\end{enumerate}
A DFA exponent greater than 0.5 indicates persistent long-range correlations. 

\textbf{Power Spectral Density (PSD).}
PSD measures how signal variance (power) distributes over frequency components, providing insight into the signal's correlation structure~\cite{heneghan2000establishing,martin2001noise}. For signals exhibiting long-range dependencies, PSD typically decays according to a power-law relationship:
\begin{equation}
S(f) \propto \frac{1}{f^\beta},
\end{equation}
where $S(f)$ denotes power at frequency $f$, and $\beta$ characterizes the correlation strength. Higher $\beta$ values indicate stronger long-range correlations. 

\textbf{Autocorrelation Function.}
The autocorrelation function quantifies the similarity of a signal with a delayed version of itself, describing how quickly correlations decay over increasing lag~\cite{durbin1950testing,box2015time,sokal1978spatial}. Signals exhibiting long-range dependency typically display slow, power-law decay in autocorrelation.

Table~\ref{tab:metrics} summarizes the average metrics obtained for each attention module:

\begin{table}[ht]
    \centering
    \small
    \begin{tabular}{lcc}
        \toprule
        \textbf{Attention Module} & \textbf{Hurst Exponent} & \textbf{DFA Exponent} \\
        \midrule
        \textit{att1} (Early Encoder)  & 0.628  & 0.459 \\
        \textit{att3} (Bottom Encoder)  & 0.594  & 0.538 \\
        \textit{att4} (Bottom Decoder)  & 0.597  & 0.501 \\
        \textit{att6} (Top Decoder)     & 0.694  & 0.657 \\
        \bottomrule
    \end{tabular}
        \caption{Average Hurst and DFA Exponents for Different Attention Modules}
    \label{tab:metrics}
\end{table}

As shown in Table~\ref{tab:metrics}, the early encoder layer (\textit{att1}) exhibits a Hurst exponent of 0.628, indicating a moderate level of persistence in the feature maps at the initial stage of encoding. However, the corresponding DFA exponent is slightly lower (0.459), suggesting that while there is some global structure, local fluctuations are still prominent.

At the bottom of the encoder (\textit{att3}), the Hurst exponent decreases modestly to 0.594 while the DFA exponent increases to 0.5379. This shift implies that the deepest encoder layer integrates more contextual information, leading to stronger long-range correlations, albeit still close to the 0.5 threshold.

In the decoder, the early stage (\textit{att4}) shows similar behavior to the bottom encoder, with a Hurst exponent of 0.597 and a DFA exponent of 0.501. These values indicate that the decoder, at this stage, continues to preserve mid-range dependencies without fully emphasizing global coherence.

\begin{figure}[ht]
    \centering
    \includegraphics[width=\linewidth]{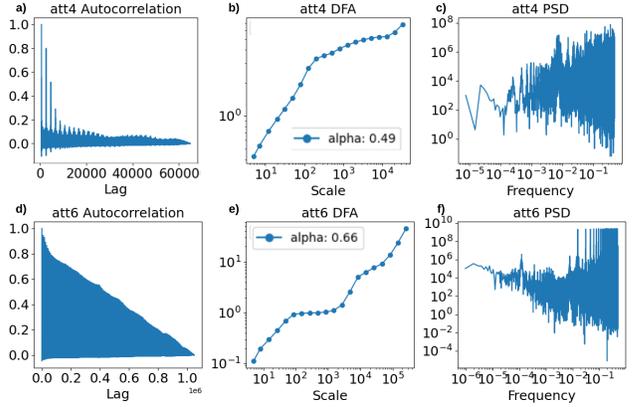}
    \caption{\textbf{Long-Range Dependency Analysis in the Decoder.} (a) Autocorrelation plot for the bottom decoder layer (\textit{att4}) demonstrating the decay of correlation. (b) DFA plot for \textit{att4}, with the slope reflecting the scaling exponent. (c) PSD plot for \textit{att4} showing frequency-domain characteristics. (d) Summary visualization of the Hurst exponent across decoder layers. (e) Summary visualization of the DFA exponent across decoder layers. (f) Combined metric overview for decoder long-range dependencies, highlighting the high persistence in the top decoder layer (\textit{att6}).}
    \label{fig:decoder}
\end{figure}

Most notably, the top decoder layer (\textit{att6}) exhibits the highest values among all modules, with a Hurst exponent of 0.694 and a DFA exponent of 0.657. These results suggest that the final decoder integrates a broader context across the entire image, thereby capturing stronger long-range dependencies. This is further illustrated in Figure~\ref{fig:decoder}, which shows the evolution of the feature correlation structure in the decoder stages.

Figures~\ref{fig:encoder} and~\ref{fig:decoder} provide visual representations of the long-range dependency characteristics in the encoder and decoder, respectively. The encoder (Figure~\ref{fig:encoder}) illustrates that although early layers capture a significant amount of local detail, the global structure is enhanced deeper in the network. On the other hand, the decoder (Figure~\ref{fig:decoder}) reveals that the top-most layers effectively synthesize this local information to generate more globally coherent feature maps.

Overall, the results demonstrate that MaskAttn-UNet progressively improves its long-range dependency capture ability throughout the network, with the top decoder layer (\textit{att6}) achieving the strongest persistence. This capability is critical for segmentation tasks, as it allows the model to integrate contextual cues from distant regions while preserving fine-grained details.

\end{document}